# A Deep Learning Approach for Blind Drift Calibration of Sensor Networks

Yuzhi Wang, *Student Member, IEEE*, Anqi Yang, Xiaoming Chen, *Member, IEEE*, Pengjun Wang, Yu Wang, *Senior Member, IEEE*, and Huazhong Yang, *Senior Member, IEEE*

*Abstract*—Temporal drift of sensory data is a severe problem impacting the data quality of wireless sensor networks (WSNs). With the proliferation of large-scale and long-term WSNs, it is becoming more important to calibrate sensors when the ground truth is unavailable. This problem is called "blind calibration". In this paper, we propose a novel deep learning method named projection-recovery network (PRNet) to blindly calibrate sensor measurements online. The PRNet first projects the drifted data to a feature space, and uses a powerful deep convolutional neural network to recover the estimated drift-free measurements. We deploy a 24-sensor testbed and provide comprehensive empirical evidence showing that the proposed method significantly improves the sensing accuracy and drifted sensor detection. Compared with previous methods, PRNet can calibrate 2× of drifted sensors at the recovery rate of 80% under the same level of accuracy requirement. We also provide helpful insights for designing deep neural networks for sensor calibration. We hope our proposed simple and effective approach will serve as a solid baseline in blind drift calibration of sensor networks.

## I. INTRODUCTION

A WIRELESS sensor network (WSN) is composed of a group of small and inexpensive sensors with the ability of sensing, measuring, data processing, and communication. WSNs can gather information from the environment and transmit the collected data to users [1]. They have important usage in many emerging applications such as environmental monitoring [2], smart cities [3], precise agriculture [4], etc. In recent years, mature WSN technologies have made it possible to deploy large-scale WSNs at an acceptable cost. In practice, many WSNs have hundreds of sensors deployed [2], [5].

With the proliferation of large-scale and long-term WSNs, sensor drift, however, has become a serious practical problem. For example, Ni *et al.* [6] give an example of a drifted

Manuscript received April 3, 2017; revised May 6, 2017; accepted May 6, 2017. Date of publication May 12, 2017; date of current version June 12, 2017. This work was supported by the National Natural Science Foundation of China under Grant 61271269 and Grant 61321061. The associate editor coordinating the review of this paper and approving it for publication was Dr. Ashish Pandharipande. *(Corresponding author: Yuzhi Wang.)*

Y. Wang, A. Yang, P. Wang, Y. Wang, and H. Yang are with the Department of Electronic Engineering, Tsinghua University, Beijing 100084, China (e-mail: yz-wang12@mails.tsinghua.edu.cn; yang-aq14@mails.tsinghua.edu.cn; wangpj@tsinghua.edu.cn; yu-wang@tsinghua.edu.cn; yanghz@tsinghua.edu.cn).

X. Chen is with the Department of Computer Science and Engineering, University of Notre Dame, Notre Dame, IN 46556 USA (e-mail: xchen7@nd.edu).

Digital Object Identifier 10.1109/JSEN.2017.2703885

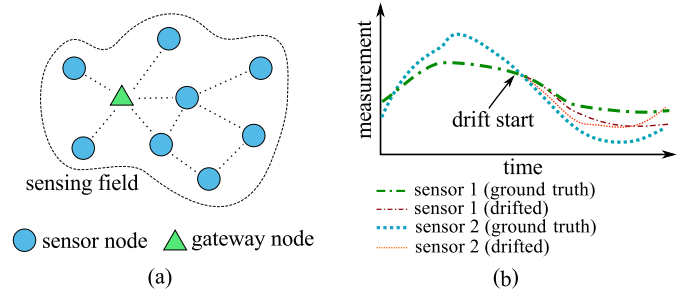

Fig. 1. Sensor networks for general monitoring: (a) typical monitoring sensor network and (b) sensor measurements and drifts.

soil $CO_2$ sensor reporting erroneous data which is about 200% of the expected ground truth. What is worse, WSNs can scale out to hundreds or even thousands of sensors which are often deployed in nearly inaccessible locations, such as wild fields and building structures. It is infeasible to unmount and re-calibrate the sensors individually. Therefore, there is an urgent need to calibrate the sensors without the ground truth data. This problem is called *blind calibration* [7].

Many general monitoring applications require blind calibration, such as environmental monitoring, structure health monitoring, precise agriculture, etc. To blindly calibrate a monitoring WSN, we must find an alternative calibration reference instead of the ground truth. However, there are two major challenges in finding an appropriate reference:

- *Lack of a prior data model*: In monitoring applications, there are many kinds of measurands, such as temperature, humidity, air quality, etc. The large amount of complicated, coupling factors make it difficult, if not impossible, to build a white-boxed data model.
- *Low-density deployment*: For a sensor network monitoring a field of interest, as depicted in Fig. 1(b), measurand data from different sensors may vary significantly. Therefore, measurements from neighbour sensors cannot be directly used as a proper reference.

In this paper, we propose a deep neural network named Projection-Recovery network (PRNet) to blindly calibrate sensor measurements online. It conquers the first challenge by learning features from sensory data as opposed to applying a prior data model, and the second challenge is overcome by utilizing the spatial and temporal correlations of data from all sensors instead of the direct equality of neighbour data. PRNet first projects the drifted measurements to a feature space to separate the mixed drift from the signal, and then recovers the drift-free measurements. Existing blind calibration methods need special assumptions, such as the linearity of the data space and the sparsity of the drift, and also use pre-defined rules for feature extraction and sensor calibration [7]–[13]. On the contrary, PRNet has less application-related assumptions and can better utilize data correlations to calibrate drifted sensors with end-to-end learning approaches. Experimental results show that PRNet brings much higher recovery rate and lower calibration error compared with existing methods.

The main contributions of this paper include:
- We propose PRNet, a novel deep neural network architecture which can automatically extract spatial and temporal features from sensory data and generate recovered drift-suppressed measurements. We also provide a data augmentation method to generate infinite samples of training data from limited sensor measurements and improve the robustness of the model. To the best of our knowledge, this is the first work that applies deep learning in sensor data calibration.
- We explore the influence of network architecture and parameter selection on the calibration accuracy, and provide comprehensive insights in designing efficient deep neural networks for spatial-temporal data processing.
- Both simulated and real-world testbed datasets are used to evaluate PRNet. Experimental results show that, compared with the existing SPSR-TSBL (subspace projection and sparse recovery with temporal correlated sparse Bayesian learning) method, PRNet can calibrate two times of drifted sensors at the recovery rate of 80% with the same level of accuracy. More benchmarks on generalization ability show PRNet can calibrate different types of drifts under noisy measurements.

The rest of this paper is organized as follows. We first formulate the blind calibration problem and review related work in Section II. Next we describe the architecture and the training method of PRNet in Section III. In Sections IV and V, we benchmark the performance of PRNet on both testbed and simulated datasets, and further explore how different settings can influence PRNet's performance. We discuss the interpretability of our method and give an intuitive explanation in Section VI. Finally, we conclude this work in Section VII.

## II. Preliminaries

In this section, we introduce necessary preliminaries of this work, including the problem formulation of blind calibration for sensor networks and existing work for this problem.

### A. Problem Formulation

Considering a sensor network deployed in a field of interest, we assume that the measurand signal is continuous within the sensing space, and the measurements collected from the sensors are spatially and temporally discrete samples of the signal field, as shown in Fig 2.

Let $N$ denote the number of sensors, and $x_{i,t}$ denote the ground truth signal value at time epoch $t$ in the position of Sensor-$i$. Ideally, Sensor-$i$ should report $x_{i,t}$ as its measurement. However, due to the existence of sensor drift and measurement noise, Sensor-$i$, in fact reports an erroneous measurement value, denoted as $y_{i,t}$. We assume that

$$y_{i,t} = x_{i,t} + d_{i,t} + v_{i,t} \quad (1)$$

where $x_{i,t}$, $d_{i,t}$ and $v_{i,t}$ represent the ground truth signal value, the drift and noise value of Sensor-$i$ at time instant $t$, respectively. Usually, sensor drift is a long-term process and smoothly increases over time [14], [15], so its value may be at the same order of magnitude with the signal. Measurement noise, however, does not accumulate over time, and, in most cases, noise is much smaller than the signal.

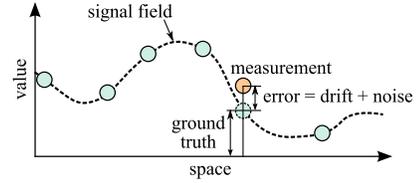

Fig. 2. Field-sample model of general monitoring sensor networks: sensor measurements are discrete samples of a continuous sensing field, the measurements have additive drift and noise

To describe the measurement model given by Eq. (1) for a sensor network, we rewrite it into a matrix notation, given by

$$\boldsymbol{Y} = \boldsymbol{X} + \boldsymbol{D} + \boldsymbol{v} \quad (2)$$

where $\boldsymbol{X}$, $\boldsymbol{Y}$, $\boldsymbol{D}$ and $\boldsymbol{v}$ represents the ground truth signal, the measured value, the sensor drift and measurement noise, respectively. Each variable, for example, $\boldsymbol{Y}$, is an $N \times T$ matrix, where $T$ is the temporal length. Therefore, each row $\boldsymbol{y}_{i,\cdot}$ is a series of measurements reported by sensor-$i$, and each column $\boldsymbol{y}_{\cdot,t}$ represents the measurements of all sensors collected at time instant $t$.

The blind calibration process is to recover the unknown ground truth $\boldsymbol{X}$ from sensor measurements $\boldsymbol{Y}$ with unknown sensor drift $\boldsymbol{D}$ and noise $\boldsymbol{v}$. Let $f_c(\cdot)$ denote a *calibration function*. Our goal is to find a function $f_c(\cdot)$ to minimize the calibration error. This optimization problem can be written as

$$\min_{f_c(\cdot)} \|f_c(\boldsymbol{Y}) - \boldsymbol{X}\| \quad (3)$$

where $\|\cdot\|$ represents a general norm operator and $\boldsymbol{X}$ is unknown. Note that this form of optimization goal is an abstract representation. The detailed measurement of the calibration error, and the constraint conditions of Eq. (3) are both specific to the application.

In general monitoring applications, the elements of $\boldsymbol{X}$ can be assumed to be correlated over row (space) and column (time). Different ways of building the calibration function result in different calibration methods.

### B. Related Work

As discussed in the above subsection, the key point of blind calibration is to find the reference signal. Many existing blind calibration methods rely on specific application features, such as a prior data model [16], [17], dense-deployment [18], or sensor mobility [19], [20].

However, for general monitoring applications, sensors are usually deployed in fixed locations at a low density. To find the reference for calibration, a reasonable assumption is that the sensory data are correlated, since they share the same feature set of the sensing field. There are mainly two kinds of calibration scheme based on this assumption: *prediction* and *subspace projection*.

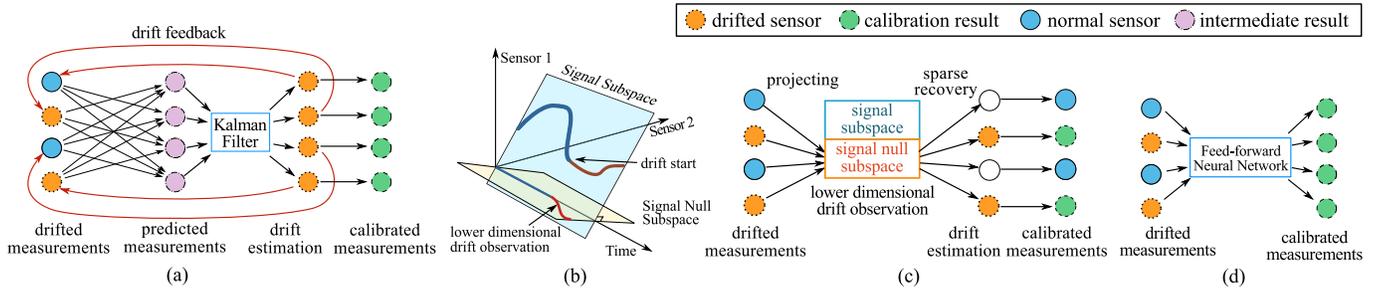

Fig. 3. Comparison of blind calibration methods: (a) illustrates the prediction-estimation-feedback loop of prediction-based methods; (b) depicts the signal subspace model and how to obtain a low-rank observation of sensor drifts; (c) shows the process of subspace-based calibration method; and (d) shows the neural network calibration method, which has the shortest pipeline among all.

The prediction-based calibration framework, which is illustrated in Fig. 3(a), is proposed by Takruri *et al.* [11], where the ground truth of a sensor is first predicted using the neighbour sensors' measurements, and then a Kalman filter (KF) is employed to track the sensor's drift. Following works apply different prediction functions in this framework, including support vector regression (SVR) [12] and Kriging interpolation [13]. However, the prediction function can only deal with drift-free measurements, so a sensor should be calibrated before it is used to predict other sensors' measurements. Therefore, there exists a prediction-calibration loop in this framework. Once the drift estimation becomes inaccurate, the feedback loop will possibly amplify the estimation error, leading to instable and erroneous calibration results.

Balzano and Nowak [7] first proposed the idea of *signal subspace* where the sensory data lie in, so a part of calibration parameters can be obtained by solving a homogeneous linear system. In this work, the calibration model contains a scaling term and an offset term. However, only the scaling term can be effectively solved, while estimating the offset term needs further assumptions. In our previous works [8], [9], we extend this idea by modeling the drift calibration problem as sparse signal recovery, and use Kalman filter or sparse Bayesian learning to estimate the sensor drift from measurements. As depicted in Fig. 3(b), the measurement space is divided into the signal subspace and its orthogonal complement, namely, the signal null subspace. The projection of sensor measurements onto the signal null subspace is fully driven by sensor drifts and noise, so this projection is a lower dimensional observation of the drift and noise. As shown in Fig. 3(c), by estimating sensor drift from the drift observation using sparse recovery methods, drifted sensors can be calibrated. Experiments [8], [9] show that the subspace methods are more stable and more accurate than the prediction methods. However, due to the systematic limitation of the under-determined calibration equations, only a portion of sensors can be calibrated.

In recent years, deep neural networks have reached the state-of-the-art performance in many applications, especially in computer vision (CV). Some applications, such as image denoising and inpaiting [21] aiming at restoring corrupted images, have some similarities with the sensor calibration problem. However, since sensory data have very different features from images, our experiments show that the networks designed for CV applications cannot work well in sensor calibration. There are also some works that apply deep learning to multivariate time-series applications. Lipton *et al.* [22] use a recurrent neural network (RNN) to detect events from segments of clinical measurements. However, training such an RNN needs a large amount of long-term time series data, which is infeasible for a specific sensor network with limited data.

TABLE I
COMPARISON OF BLIND CALIBRATION METHODS IN PERFORMANCE

| Method | Calibratable Sensors | Stability | Accuracy |
| --- | --- | --- | --- |
| Subspace | partial calibration | good | high |
| Prediction | all sensors | bad | medium |
| Proposed | all sensors | good | high |

In this paper, we propose a convolutional neural network (CNN) to blindly calibrate general monitoring sensor networks. As shown in Fig. 3(d), the proposed method directly maps the drifted measurements to drift-suppressed measurements. Similar to the subspace and prediction methods, the calibration function is learned from the sensory data. However, the subspace and prediction methods have two steps: 1) learning the subspace or prediction function, and 2) recovering sensory data using pre-defined rules. In these two steps, only the learning step can fully utilize data features. The proposed neural network, on the contrary, is an end-to-end method, where the feature learning and the drift compensation steps are modeled as different layers, which are jointly trained using sensory data. This means that the proposed method can make better use of data correlations and learn a better data model. In Table I, we give a qualitative comparison of the three blind calibration schemes.

## III. CALIBRATION WITH DEEP FULL CONVOLUTIONAL NEURAL NETWORK

CNNs [23] are widely used in image processing. To design a CNN-based method for blind calibration of sensor networks, there are two major challenges:

- Many existing works on CNNs are proposed for image processing. How do we design the network architecture for sensor drift calibration?

- Deep neural networks need to be trained with a large amount of training data, whereas the sensory data collected from a specific sensor network are limited. How to train the neural network with limited sensory data?

To solve the first issue, we extend the idea of the previous SPSR framework [9] by designing a projection-recovery CNN architecture named PRNet. The first layer projects the drifted measurements to a feature space, and this layer is trained to keep the drift features. The following recovery layers are trained to fuse the features to drift-free output. When the training converges, the network can automatically extract features from sensory data and fuse these features to drift-free measurements.

Similar to previous works [9], [12], we assume that the sensors are calibrated before deployment, so the sensory data collected within a short period after deployment can be regarded as drift-free. We propose a data augmentation method which generates training data from a relatively small dataset. Thus the second issue can be solved.

### A. Convolution on Sensory Data

Before describing the architecture of PRNet, we present our basic idea of applying convolutions to sensory data. Let $Y_{N \times T}$ denote the matrix containing sensory data collected from a sensor network, where the rows represent measurements from $N$ different sensors, and the columns are measured at $T$ different time instants. The input of a convolution layer is a 3-D tensor, denoted as $X_{(N,T,c)}$, where $N$, $T$, and $c$ represent its numbers rows, columns, and channels, respectively. Therefore, we convert the measurement matrix $Y_{N \times T}$ to a tensor $Y_{(N,T,1)}$ so that it can be fed to a convolution layer.

A convolution layer consists of several *convolution kernels*. Each of them is a filter with the size of $(k_s, k_t, c)$, which maps a patch of its input tensor to a scalar, given by

$$x_{\text{out}} = \sigma(\sum W_{(k_s,k_t,c)} \circ X_{(k_s,k_t,c)} + b) \quad (4)$$

where $W_{(k_s,k_t,c)}$ and $b$ are the parameters of the convolution kernel and the bias term respectively; $X_{(k_s,k_t,c)}$ is a patch of the input tensor; $\circ$ is the Hadamard product, and $\sigma$ represents a nonlinear *activation function*. Usually, as $c$ is decided by the input tensor, we use $k_s \times k_t$ to denote the size of a convolution kernel. As shown in Fig. 4, a convolution kernel slides over the space (row) and time (column) dimension of the input tensor, and maps a $c$-channeled input tensor to an 1-channel output tensor. Since a convolution layer contains $c_{\text{out}}$ kernels, its output tensor has $c_{\text{out}}$ channels. The rows and columns of the output tensor are decided by the padding size of the input tensor and the sliding stride of the filter.

Fig. 4 shows a basic CNN applied to sensor measurements. By properly setting the padding size on the input tensors, the numbers of rows and columns of all tensors can be kept $N$ and $T$. Because of the cascading structure of CNNs, each element of a feature tensor is a fusion of sensory data from multiple sensors measured at multiple time instants, and this patch of sensory data is named as a *receptive field*. Therefore, a CNN can utilize both spatial and temporal correlations of sensor measurements.

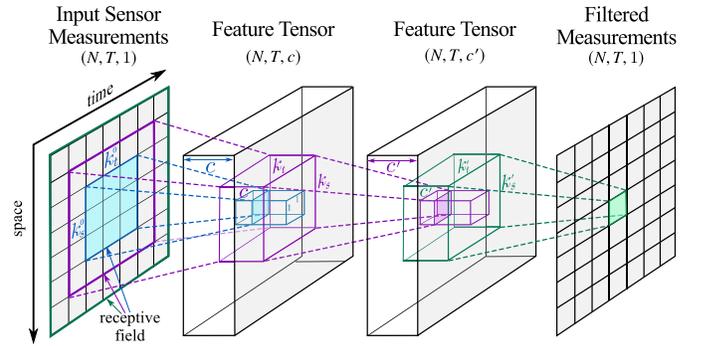

Fig. 4. Basic architecture of a CNN for sensor network data. Each convolution kernel fuses a block of its input tensor to a pixel in the next layer. Each pixel in the 2nd, 3rd and 4th layer respectively comes from a $3 \times 3$, $5 \times 5$ and $7 \times 7$ sub-block of the input, named as the receptive field.

The last convolution layer has only one convolution kernel and outputs an $(N, T, 1)$ tensor, which is of the same size as the input sensor measurements. The task of the output layer is to decode the features extracted by the previous layers and fuse them into drift-suppressed measurements.

### B. Architecture of PRNet

The architecture of PRNet is derived from the basic CNN for sensory data. We first review the key idea of the previous state-of-the-art subspace-based calibration framework. According to [8], if the drift-free measurements lie in a signal subspace, a projection matrix $P$ can be obtained, which satisfies

$$PY = P(X + D) = PD \quad (5)$$

where $X$, $D$ and $Y$ represents the ground truth signal, sensor drift and drifted measurements, respectively. This equation is the key point of the subspace method, since the projection eliminates the unknown ground truth signal, obtaining a lower-dimensional observation of the drift.

In our work, we extend this drift projection by implementing it with a $N \times \tau$ global convolution layer, where $N$ is the number of sensors and $\tau$ is a temporal window size. Recalling the convolution function of a single kernel in Eq. (4), let $w = \text{vec}(W_{(N,\tau,1)})$ and $y_\tau = \text{vec}(Y_{(N,\tau,1)})$, where $W_{(N,\tau,1)}$ is the weight tensor of the convolution kernel, and $Y_{(N,\tau,1)}$ is a temporal patch of the drifted measurement tensor, and $\text{vec}(\cdot)$ stacks the columns of a tensor to a column vector. Eq. (4) can be rewritten as

$$x^{\text{out}}_{(1,1)} = \sigma(w^T y_\tau + b). \quad (6)$$

For a convolution layer with $R$ kernels, we stack the weight vectors to a matrix and bias terms to a vector by setting $W = [w_1, w_2, \ldots, w_R]^T$ and $b = [b_1, b_2, \ldots, b_R]$. The output function of this convolution layer is

$$x^{\text{out}}_{(1,1,R)} = \sigma(W y_\tau + b). \quad (7)$$

Thus, we obtain the convolution-based projection, which is a natural extension to the linear projection $PY$ with two advantages. First, the convolution is applied to multiple measurement vectors, so temporal correlations of the sensing signal can be utilized. Second, with the nonlinear activation function, this projection can be applied to measurements from nonlinear sensing fields. If $\tau$ is set to 1 and the activation function is linear, this convolution projection is equivalent to the linear projection used by subspace methods.

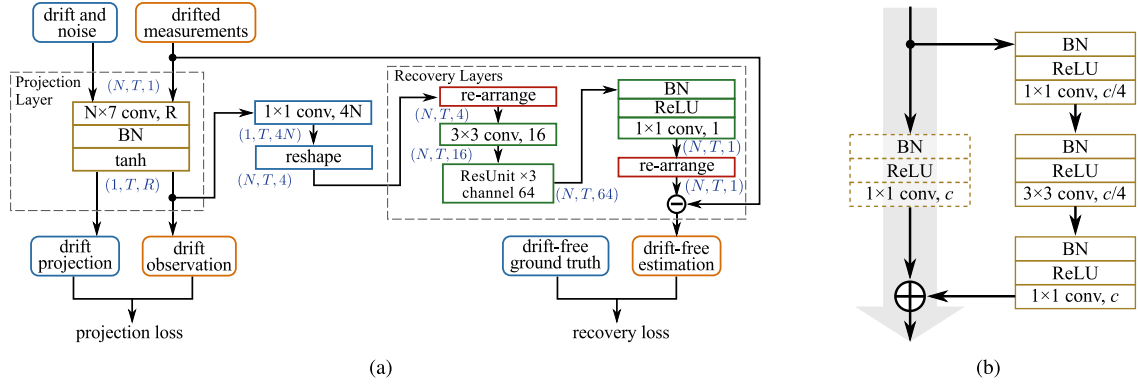

Fig. 5. (a) Overall architecture of PRNet. The input/output feature map size is annotated beside each layer. (b) Structure of a ResUnit [24]. The dashed $1 \times 1$ convolutional layer only exists in the first ResUnit to increase the feature map channels.

Fig. 5(a) illustrates the overall architecture of the proposed PRNet. The projection layer is a convolution layer with kernel size $N \times 7$, and the activation function is $\tanh(\cdot)$. The hyperparameter $R$, which is the number of convolution kernels and the dimension of the projection space, should be decided based on the number of sensors. In practice, a number around $2N$ is an appropriate choice. The term BN refers to Batch Normalization [25], which helps accelerate the training.

After the drift observation is obtained, the following layers estimate the drift. As the output size of the projection layer is $(1, T, R)$, we first up-sample and reshape it to an $(N, T, 4)$ tensor with a convolution and reshaping layer, followed by a special re-arrangement layer to put the measurements from neighbour sensors into adjacent rows, which will be discussed in Section III-C.

Next, several convolution layers extract and fuse features from the projected tensor. The architecture of these recovery layers is derived from ResNet [24], [26], which is a state-of-the-art CNN architecture widely used in CV applications. The basic component of the recovery layers is a *ResUnit*, depicted in Fig. 5(b). Each ResUnit has two branches. The main branch is an identity shortcut, which directly passes the input to the output, and the auxiliary branch contains convolution layers. The outputs of the two branches are added before being fed to the next layer. The first ResUnit, as the channel size increases from 16 to 64, includes an $1 \times 1$ convolution layer in the main branch to ensure the feature maps in its two branches have the same number of channels. Some recent works [27], [28] found that this special architecture has better representational ability and is easier to train than conventional CNN architectures.

The output of the last convolution layer is the drift estimation. By subtracting the estimated drift from the drifted measurements, we obtain the estimation of drift-free measurements.

### C. Sensor Re-Arrangement

In this subsection, we focus on the input and output data of the recovery layers. As discussed before, the recovery layers estimate the drifts by utilizing the data correlation within the receptive field of their input data. However, the sizes of their receptive fields are limited. Therefore, it is required that the adjacent rows and columns of the input data should be as correlated as possible. The columns of the input data represent time instants, which are naturally ordered and the adjacent data are naturally correlated. However, the row order of the input data depends on the numbering of sensors. Therefore, we need to re-arrange to rows of the input data in a proper order.

We model the sensor re-arrangement operation as a simple matrix multiplication. For a matrix $M_{(K,N)}$, it is a re-arrangement matrix if and only if the following conditions hold:

$$\begin{cases} K \geq N \\ m_{i,j} \in \{0, 1\} \quad \forall m_{i,j} \in M_{(K,N)} \\ \sum M_{i,\cdot} = 1 \quad \forall i \in \{1, 2, \ldots K\} \\ \sum M_{\cdot,j} \geq 1 \quad \forall j \in \{1, 2, \ldots N\} \end{cases} \quad (8)$$

Eq. (8) means that a re-arrangement matrix is composed of 0 and 1, each row of which only contains one 1 and each column of which contains at least one 1s. Given a measurement matrix $X_{(N,T)}$, by left multiplying $X_{(N,T)}$ by $M_{(K,N)}$, we obtain a re-arranged measurement matrix, given by

$$X^R_{(K,T)} = M_{(K,N)} X_{(N,T)}. \quad (9)$$

As $K$ can be larger than $N$, each row of $X_{(N,T)}$ can appear multiple times in $X^R_{(K,T)}$.

By modeling the re-arrangement operation as matrix multiplication, it is easy to be implemented as a re-arrangement layer in a neural network. We put the re-arrangement layer before the recovery layers, as shown in Fig. 5(a). In addition, after the drift estimation is obtained, we need to inversely arrange the rows of the drift-estimation matrix to match the original sensor order. This operation can also be easily implemented by left multiplying the drift-estimation matrix by an inverse-arrangement matrix $M^R_{(N,K)}$, which is the row-normalized transpose of the re-arrangement matrix, given by:

$$M^{R0}_{(N,K)} = M^T_{(K,N)} \quad (10)$$

$$M^R_{i,\cdot} = M^{R0}_{i,\cdot} / \|M^{R0}_{i,\cdot}\|_0. \quad (11)$$

Next, we discuss how to decide the order of the re-arranged matrix. Usually, we can have some prior assumptions on the correlations of different sensors. Without loss of generality, we assume that neighbour sensors are more correlated. In other words, the correlation between two sensors depends on their distance. We denote the distance between sensor-$i$ and sensor-$j$ as $d(i, j)$, and let $s[k]$ be the sensor number corresponding to the $k$-th row of the re-arranged matrix. To maximize the local correlation of the re-arranged matrix, we need to give an optimal mapping $s[\cdot]$ to minimize the maximum sensor distance, and every sensor should be included in this mapping, written as

$$\min \max_{i=1}^{K} d(s[i], s[i+1]) \tag{12}$$

$$\text{s.t.} \bigcup_{i=1}^{K} s[i] = \{1, 2, \ldots, N\}. \tag{13}$$

The optimal solution of Eq. (12) can be obtained with the following steps:
1. Generate a minimum spanning tree (MST) on the sensor graph;
2. Duplicate every edge of the MST, obtaining a Eulerian graph;
3. Set $s[\cdot]$ by traversing over a Eulerian circuit over the Eulerian graph.

Besides the optimal mapping, an approximation can also be employed, since the convolution kernels in PRNet have the ability to utilize sensor data in its receptive field. In our experiments, for simplicity, we use a greedy nearest neighbour algorithm to generate $s[\cdot]$. We first set $s[1]$ to 1, and choose the nearest non-visited neighbour sensor for the next step until all sensors are visited.

As long as $s[\cdot]$ is obtained, $M_{(K,N)}$ can be calculated by setting $M_{i,s[i]}$ to 1 for all $i \leq N$, and other elements to zero.

### D. Training Data Generation

As sensor networks deployed in different sensing fields vary a lot in data features, the calibration model for a specified sensing field must be trained using the sensory data collected from the very same field. To train PRNet, pairs of drifted and drift-free measurements are required, but for a deployed sensor network, both the drift-free signal and the drift are unknown. Moreover, the amount of sensory data from one single sensor network is limited, while a neural network must be trained with a large amount of data. Therefore, we propose a data synthesis and augmentation method to build the training dataset.

We assume that the sensors are calibrated before deployment, so the sensory data collected within a short period after deployment should contain negligible drift and carry the features of the sensing field. On the contrary, sensor drift and noise, are usually caused by errors and non-ideal factors of sensor hardware [6]. Therefore, we can use drift-free measurements and simulated sensor drift to synthesize drifted measurements. What is more, as infinite samples of sensor drift can be generated according to any possible model, the amount of training data also gets increased.

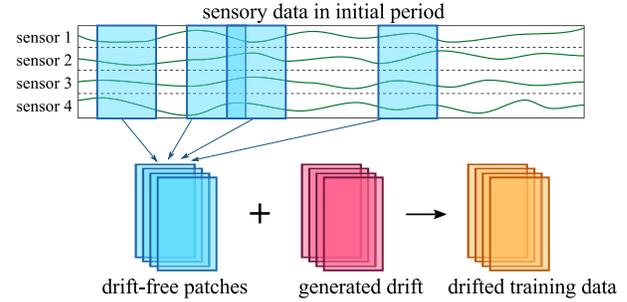

Fig. 6. Demonstration of the data augmentation process: randomly picking patches from the training dataset and adding simulated random drift to obtain training data.

Since the temporal sizes of convolution kernels in PRNet are limited, we do not need to use all the training data for each iteration. Instead, we apply random-cropping to generate small patches of measurement data. This is also widely used in other research areas such as time-series bootstrapping [29], [30]. We denote the sensory data of the initial period as $X^I$, which is an $N \times T_I$ matrix, where $T_I$ is the length of the initial period. In each iteration, we randomly crop a small $N \times T_P$ patch from $X^I$ denoted as $X^P$. Therefore, we can generate $T_I - T_P$ patches from the initial sensory data by

$$\{X^P\} = \{X^I_{\cdot, \tau:\tau+T_P} | \forall \tau \in [1 \ldots T_I - T_P]\} \tag{14}$$

where $\{X^P\}$ stands for the set of patches, and $X^I_{\cdot, \tau:\tau+T_P}$ is an $N \times T_P$ sub-block of $X^I$ from column $\tau$ to $\tau + T_P$. Thus, we obtain $T_I - T_P$ segments of sensory data which carry the features of the sensing field. Besides, by cropping the measurements to small patches, the randomness and diversity of the training set get increased, which can help avoid overfitting.

The selection of $T_P$ depends on the receptive field size over the time dimension of the neural network, denoted as $R_t$. If $T_P$ is smaller than $R_t$, some layers of the neural network will not be able to obtain enough training data. Our experiment shows that a small $T_P$ which is slightly larger than $R_t$ is appropriate.

Fig. 6 illustrates the augmentation process. We randomly crop a drift-free patch $X^P$, and randomly generate $N$ drift and noise samples, denoted as $D^P$ and $v^P$. Thus, a patch of drifted measurements can be generated by adding the drift and noise to the drift-free patch, denoted as

$$Y^P = X^P + D^P + v^P. \tag{15}$$

In practice, the drift and noise generation model should be designed based on the sensor type and application requirements. If required, more kinds of corruptions can be added to the measurements, such as random bias, temporary data loss, etc. In this paper, to demonstrate a general calibration process, without loss of generality, noise is modeled as white Gaussian and the sensor drift is modeled as a random walk process. The sensor noise is generated by

$$v_{i,t} \sim \mathcal{N}(0, \sigma_n^2). \tag{16}$$

We assume that the drift of different sensors are independent, and drift increments at different time instants are i.i.d. and Gaussian. This can be written as

$$d_{i,t} = d_{i,t-1} + \delta_{i,t}, \quad \delta_{i,t} \sim \mathcal{N}(0, \sigma_{i,t}^2) \tag{17}$$

where $d_{i,t}$ is the drift value, and $\delta_{i,t}$ is the increment of Sensor-$i$'s drift at time instant $t$. Therefore, the sensor drift is the accumulation of a series of Gaussian increments, which is still Gaussian [31], given by

$$d_{i,t} \sim \mathcal{N}(0, \sum_{\tau=0}^{t} \sigma_{i,\tau}^2). \tag{18}$$

Note that Eq. (18) is a time-irrelevant model representing the prior distribution of sensor drift at a single arbitrary time instant. Although it indicates that the expectation of sensor drift is zero, it does not mean that a specific series of sensor drift is zero-mean as Eq. (18) ignores the temporal correlation of drift values.

Combining Eqs. (18) and (17), we generate drift patches by

$$d_{i,0} = \mu_i + \beta, \quad \mu_i \sim \mathcal{N}(0, \sigma_0^2), \quad \beta \sim \mathcal{N}(0, \sigma_b^2) \tag{19}$$
$$d_{i,t} = d_{i,t-1} + \delta_{i,t}, \quad \delta_{i,t} \sim \mathcal{N}(0, \sigma_d^2). \tag{20}$$

Eq. (19) describes the generation of the start value of the drift patch. Compared with Eq. (18), in addition to the Gaussian variable $\mu_i$ which varies among sensors, we also add a global Gaussian bias value $\beta$. This is to add a small nonzero offset to the drift of sensors within a single patch. Eq. (20) is simplified from Eq. (17), where $\sigma_d^2$, the variance of drift increment, is a constant value. Another issue is that in many cases, not all sensors in a network are drifted. Therefore, we randomly set $d_{i,\cdot}$ to zero at the probability of 0.5.

The selection of the drift level parameters $\sigma_0$, $\sigma_b$ and $\sigma_d$ depends on the application requirements. As PRNet learns features of sensory data from drifted patches, it is best that the simulated drift is slightly greater than the real possible drift. Considering that the variance of $d_{i,0}$ is $\sigma_0^2 + \sigma_d^2$, according to the three-sigma rule, in our experiment, we set $3\sqrt{\sigma_0^2 + \sigma_d^2}$ to 60% of the dynamic range of the sensor measurements.

The algorithm for training patch generation is listed in Algorithm 1. In lines 1-10, we first generate an $N \times T_P$ noise patch and drift patch according to Eq. (16), Eq. (19) and Eq. (20), and then randomly crop a drift-free patch from the initial sensor measurements according to Eq. (14) in lines 13 and 14. Finally, corresponding to Eq. (15), by adding the generated drift and noise to drift-free measurements in line 15, we obtain the drifted measurements. As we can generate $T_I - T_P$ different patches of drift-free measurements and infinite drift samples, we manage to generate a large amount of data to train PRNet.

*E. Training*

The training process of a neural network is to minimize a loss function with respect to input data by adjusting the network parameters. The loss function of PRNet includes the projection loss and the recovery loss, denoted as

$$L_{PR} = L_P + L_R. \tag{21}$$

Recalling Eq. (5) that the key function of the projection layer is to obtain a drift observation from drifted measurements

---

**Algorithm 1**: Generating a Patch for Training

**Input**　: $X_{N \times T_I}^{\mathrm{I}}$: initial sensor measurements,
　　　　　$T_P$: patch length,
　　　　　$\sigma_0, \sigma_b, \sigma_d, \sigma_n$: drift and noise parameters
**Output**: $Y^{\mathrm{p}}$: a patch of drifted measurements,
　　　　　$X^{\mathrm{p}}$: a patch of drift-free measurements

1　$\mathbf{v} \leftarrow (v_{ij})_{N \times T_P}$ with $v_{ij}$ sampled from $\mathcal{N}(0, \sigma_n^2)$ ;
2　$\beta \leftarrow$ random number from $\mathcal{N}(0, \sigma_b^2)$;
3　**for** $i \leftarrow 0$ **to** $N-1$ **do**
4　　　$r \leftarrow$ random number from $U(0,1)$ ;
　　　　// each sensor has the probability of 0.5 to be drifted
5　　　**if** $r \leq 0.5$ **then**
6　　　　　$\mu_i \leftarrow$ random number from $\mathcal{N}(0, \sigma_0^2)$ ;
7　　　　　$d_{i,0} \leftarrow \mu_i + \beta$ ;
8　　　　　**for** $t \leftarrow 1$ **to** $T_P - 1$ **do**
9　　　　　　　$\delta_{i,t} \leftarrow$ random number from $\mathcal{N}(0, \sigma_d^2)$ ;
10　　　　　　$d_{i,t} \leftarrow d_{i,t-1} + \delta_{i,t}$;
11　　　**else**
12　　　　　$d_{i,\cdot} = 0$;
13　$\tau \leftarrow \lfloor$random number from $U(0, T_I - T_P)\rfloor$;
14　$X^{\mathrm{p}} \leftarrow X_{\cdot,\tau:\tau+T_P}^{\mathrm{I}}$ ;　　// retrieve random patch
15　$Y^{\mathrm{p}} \leftarrow X^{\mathrm{p}} + \mathbf{d} + \mathbf{v}$ ;

---

to approximate the projected ground truth drift. Therefore, the projection loss is designed as

$$L_P = \frac{1}{2|\mathbb{D}|NT_P} \sum_{i \in \mathbb{D}} \|f_p(Y_i^P) - f_p(Y_i^P - X_i^P))\|_F^2 \tag{22}$$

where $X^P$ and $Y^P$ are patches of drift-free and drifted measurements; $f_p(\cdot)$ represents the function of the projection layer, and $\mathbb{D}$ represents the training dataset.

For the recovery loss, we simply use the mean square error (MSE) between the calibrated measurements and the ground truth signal:

$$L_R = \frac{1}{2|\mathbb{D}|NT_P} \sum_{i \in \mathbb{D}} \|f_{PR}(Y_i^P) - X_i^P\|_F^2 \tag{23}$$

where $f_{PR}(\cdot)$ denotes the overall forward function of PRNet. Note that the parameters of the projection layer are also included in this loss, so there are two optimization objectives for the projection function.

Eq. (23) is a concrete form of Eq. (3), where our calibration function is built by designing the architecture and training the parameters of PRNet, and the norm of calibration error is MSE.

We use the Adam optimizer [32] to minimize the loss function. The training dataset is divided into mini-batches which are used to train PRNet. When the optimization converges, PRNet will have learned to extract spatial and temporal features of the sensing field and to suppress sensor drift.

Another issue about the training dataset is the parameter selection. We apply a simple curriculum learning [33] strategy on selecting the drift emulation parameters $\sigma_0$, $\sigma_b$, $\sigma_d$ and $\sigma_n$.

Because we need the trained neural network to calibrate sensor drift on different levels, $\sigma_0$ and $\sigma_b$ should be large enough so that the augmented training data can cover more drift levels. However, we found that directly training the neural network with large drift and noise from scratch may not converge. Therefore, we first pre-train the neural network with small drift and noise, then fine-tune the trained neural network with larger drift and noise. Thus, this neural network can deal with different levels of sensor drifts.

## IV. Performance Evaluation

In this section, we use a real-world sensor dataset to evaluate the proposed method and compare our method with two existing blind calibration methods.

### A. Datasets

Although there are many WSN projects deployed and running, and some of which have open datasets, we do not know whether those sensors are drifted and how accurate the measurements are. After calibrating these measurements, the correctness of calibration results is unknown either. Therefore, we set up a testbed which has multiple redundant sensors in every position to ensure accurate measurements, and use simulated drift and noise to benchmark calibration algorithms. Besides, as the scale of testbed is limited, we also simulate a nonlinear sensing field to generate a more challenging dataset.

*1) Testbed Dataset:* We use the same testbed described in our previous work [9] deployed in our lab to build the dataset. The testbed consists of 6 sensing units deployed in different locations in our lab. Each sensing unit has 4 sensors measuring temperature at the same location, including a commercial thermometer (type WSB-1-H2, prices at $50 each) and 3 cheap temperature sensors (type DS18B20, prices at $0.5 each). The DS18B20 sensors report their measurements every 30 seconds, and the thermometers' measurements are collected every 5 minutes.

We use the collected data from March 1 to April 24, 2016, then re-sample it to a 3-minute interval, and drop the corrupted samples. Hence, the dataset contains measurements from 24 sensors, and each sensor has 25 935 samples. For each DS18B20 sensor, we calculate its offset to its corresponding thermometer. The offset of different sensors varies from $-0.8\,°C$ to $0.5\,°C$, but for each sensor, its offset variation is within $\pm 0.1\,°C$ over time. Therefore, we use the mean offset to calibrate each DS18B20 sensor, which has the measurement error of $\pm 0.1\,°C$ after calibration, and the calibrated dataset can be considered drift-free.

We plot the first 10 000 samples of testbed data from 4 selected sensors in Fig. 7(a). We can see that the measurements from sensors in the same sensing unit are almost identical, while different sensing units report different measurements.

*2) Simulated Dataset:* The real-world testbed dataset is collected by 24 sensors deployed in 6 locations in a room, which can not provide much inter-sensor variation. To fully benchmark the performance of PRNet, we simulate a more challenging dataset for calibration algorithms.

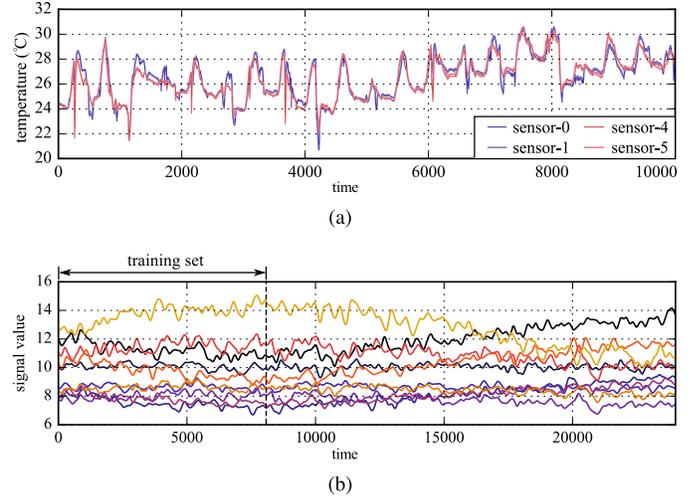

Fig. 7. Samples of datasets for evaluation: (a) measurements from testbed dataset collected from 4 sensors in 2 sensing units, where the measurements are grouped in 2 lines and (b) ground truth signals from 10 randomly selected sensors in the simulated dataset, where each sensor has different measurements and trends, and the value range in the training set and the rest dataset are also quite different.

We simulate a circular sensing field with radius 10, where 50 sensors are deployed in random locations. The drift-free measurement of a sensor is a nonlinear combination of 20 *signal sources*, given by

$$x_{i,t} = \left(\sum_{j=1}^{20} a_{j,i} s_{j,t}\right)^{1/2} + \sqrt{(\dot{a}_i \dot{s}_{i,t})(\ddot{a}_i \ddot{s}_{i,t})} \quad \forall i \in \{1, 2, \ldots, 50\} \tag{24}$$

where $x_{i,t}$ is the drift-free measurement of sensor-$i$ at time instant $t$, and $s_{j,t}$ represents the signal value of source-$j$. The special terms $\dot{s}_{i,t}$ and $\ddot{s}_{i,t}$ represent the nearest two signal sources of sensor-$i$. The combination coefficient $a_{j,i}$ is determined by the distance between a sensor and a source, given by

$$a_{j,i} = (\Delta_{j,i} + 1)^{-1.5} \tag{25}$$

where $\Delta_{j,i}$ represents the distance. For each signal source, we independently simulate 24 000 samples of its signal values with a lowpass-filtered ARMA (Autoregressive moving average) process, plus a random trend signal. Using Eqs. (25) and (24), we obtain the drift-free measurements of 50 sensors, each of which has 24 000 samples. Therefore, the measurements of each simulated sensor is different and non-stationary, but still have nonlinear correlation. We randomly selected 10 sensors and plot the simulated measurements in Fig. 7(b).

### B. Training Settings

In both the testbed and simulated datasets, we use the first 8000 samples of each sensor to build the training dataset. The augmentation patch length $T_p$ is set to 20, and the mini-batch size is set to 64. For the testbed dataset, we set the dimension of the projection space to 64, and for the simulated dataset, we set it to 128. The weight parameters are initialized using the initialization method proposed by He *et al.* [34], and the bias parameters are initialized to zeros. As the sensor measurement matrix is already ordered in sensing units, we use an identity matrix to bypass the re-arrangement. For the simulated dataset, we obtain the re-arrangement matrix using the nearest-neighbour algorithm.

During the pre-training process, the drift generation parameters $\sigma_0$, $\sigma_b$ and $\sigma_d$ are set to 0.5, 0.2 and 0.02, respectively, and the noise parameter $\sigma_n$ is set to zero. The learning rate is set to $1 \times 10^{-3}$, and then updated to $1 \times 10^{-4}$ at the $10\,000^{\text{th}}$ iteration, and finally updated to $1 \times 10^{-5}$ at the $40\,000^{\text{th}}$ iteration. The pre-training stops at the $50\,000^{\text{th}}$ iteration.

In the fine-tuning process, we set $\sigma_0$ to 1.5, $\sigma_b$ to 0.5 and $\sigma_d$ to 0.03, and add noise with $\sigma_n$ set to 0.5. The learning rate starts at $2 \times 10^{-4}$, and then updates to $1 \times 10^{-4}$ at the $10\,000^{\text{th}}$ iteration and decreases to $1 \times 10^{-5}$ at the $20\,000^{\text{th}}$ iteration. It takes 30 000 iterations for fine-tuning.

In the rest of this paper, if not specified, we keep these experiment settings unchanged.

### C. Comparison With Existing Methods

We compare the calibration performance of PRNet with two representative existing calibration methods, including
- SPSR-TSBL: a subspace-based method proposed in [9], which projects drifted measurements to a lower-dimensional drift-observation subspace, and then estimates drift values using the T-SBL algorithm;
- SVR-KF: a prediction-based drift calibration algorithm proposed in [12], which uses SVR to predict sensor measurements and a Kalman filter to smooth the estimated drift.

We use two indicators to evaluate the performance of calibration algorithms. One is the rooted mean square error (RMSE) between the ground truth and the calibrated measurements, which indicates the calibration accuracy. It is defined as:

$$\text{RMSE}(X, \hat{X}) \stackrel{\text{def}}{=} \sqrt{\frac{\|X - \hat{X}\|_F^2}{NT}}$$

where $X$ represents the ground truth and $\hat{X}$ is the calibrated measurements. The other is the *recovery rate*, which measures the method's ability to detect and identify the drifted sensors when a part of sensors are drifted. For a sensor network with $m$ drifted sensors, by calibrating their measurements, we can estimate each sensor's drift. We select $m$ sensors with the largest estimated drift as the guess of drifted sensors. The drift recovery is *successful* only if the guessed $m$ sensors are exactly those drifted ones. In our experiments, we can simulate a number of drift samples on different sensors, so we can run many times of the drifted sensor detection process, and the ratio of successful trials among all experiments is defined as the recovery rate.

The drift is simulated with a random walk process, given by

$$d_{i,0} = 0$$
$$d_{i,t} = d_{i,t-1} + \delta_{i,t}, \quad \delta_{i,t} \sim \mathcal{N}(0, \tilde{\sigma}_d^2). \quad (26)$$

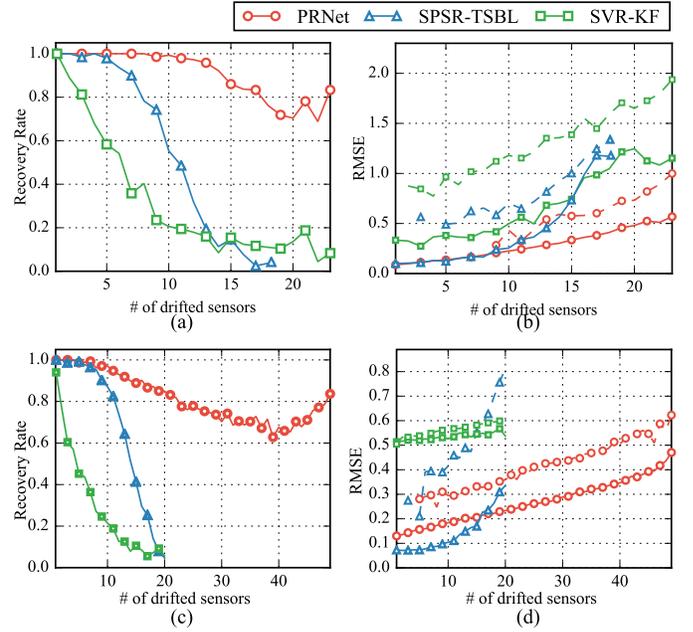

Fig. 8. Recovery rates and calibration RMSE of PRNet, SPSR-TSBL and SVR-KF: (a) and (b) on testbed dataset; (c) and (d) on simulated dataset; in (b) and (d), solid line represents calibration RMSEs of successful recoveries and dashed line represents failure cases.

For the testbed dataset, we set the number of drifted sensors $m$ to vary from 1 to 23, and for the simulated dataset, $m$ varies from 1 to 49. At each drift number $m$, we randomly choose $m \times (N - m + 1)$ different combinations of drifted sensors with independently generated drift samples. $N$ is the number of all sensors, which is 24 in the testbed dataset and 50 in the simulated dataset. The drift level parameter $\tilde{\sigma}_d$ is set to 0.02. For the T-SBL algorithm, according to [9], we set the block size $L$ to 5. The dimension of the signal subspaces estimated by SPSR-TSBL are 5 in the testbed dataset and 17 in the simulated dataset, which means the theoretical limits of drift sensors for the SPSR framework are 18 and 32 [8]. For SVR-KF, we also use the first 8000 samples of measurements to train the SVR prediction model.

Figs. 8(a) and 8(c) plot the recovery rates of the three methods at different numbers of drifted sensors on testbed and simulated dataset, respectively. The proposed PRNet achieves much higher recovery rates than the other two. On the testbed dataset, PRNet's recovery rate is always higher than 0.6. For SPSR-TSBL, when more than 10 out of 24 sensors are drifted, its recover rate is below 0.6. The SVR-KF method has the lowest recovery rate among all. When more than 3 sensors are drifted, its recovery rate goes below 0.6. Similar results can be observed on the simulated dataset.

Figs. 8(b) and 8(d) compare the calibration RMSEs on the testbed and simulated dataset respectively. The solid lines represent the RMSEs of the successful recoveries, and the dashed lines are the RMSEs of the failed recoveries. For the testbed dataset, PRNet has the lowest RMSE among all. When fewer than 10 sensors are drifted, the RMSE of SPSR-TSBL is almost the same as that of PRNet, but it increases rapidly when the number of drifted sensors gets over 10. SVR-KF has the largest RMSE in the test. For the simulated dataset, however, when the number of drifted sensors is below 15, SPSR-TSBL has the lowest RMSE on successful recoveries, but in the failure cases, PRNet has lower RMSE than SPSR-TSBL. When more than 15 sensors are drifted, PRNet has the lowest RMSE on both successful and failed recoveries.

In this experiment, PRNet shows superior performance in drifted sensor detection, where its recovery rate is much higher than the other two. The SVR-KF method does not have the ability to detect sensor drift, so it has the lowest recovery rate, and its way of predicting sensor measurements also limits its calibration accuracy. Due to the theoretical limitations and the sparsity assumption of SPSR framework, it has an obvious performance drop in both recoverability and accuracy when many sensors are drifted. Another weakness of the SPSR framework is that when it fails to correctly detect the drifted sensors, the overall calibration error increases sharply. This is also caused by its sparsity assumption. On the contrary, the proposed PRNet has neither a limitation nor a sparsity assumption on the number of drifted sensors. Instead, it is trained to calibrate any number of drifted sensors, so its calibration performance does not drop a lot when many sensors are drifted.

### D. Generalization Ability: Different Types of Drift and Noise Tolerance

In this experiment we test the generalization ability of PRNet. As we generate sensor drifts using a random-walk model during training, we need to test PRNet's performance under other types of drift. We also test its calibration performance on noisy data. The simulated dataset is used in these two tests.

In addition to the random-walk drift, we also simulate other four types of drift: *bilateral linear* drift, *positive linear* drift, *positive sqrt* drift and *sine* drift. To simulate these kinds of drift, for each sensor, we first generate a random number called the *end value* as the largest drift value it can reach. Next, we generate linear drift by

$$d_{i,t} = e_i \times t/T \quad (27)$$

and sqrt drift by

$$d_{i,t} = e_i \times \sqrt{t/T} \quad (28)$$

and sine drift by

$$d_{i,t} = e_i \times \sin(r_i \pi t/T) \quad (29)$$

where $d_{i,t}$ is the drift value of Sensor-$i$ at time instant $t$; $e_i$ is the end value; $r_i$ is a random number sampled from $U(3, 4)$ and $T$ is the total time length of the simulation. For the bilateral linear drift model, its end value is sampled from a uniform distribution $U(-S, S)$, whereas the end values for positive linear drift, positive sqrt drift and sine drift are sampled from $U(0, S)$, where $S$ is the parameter to control the drift level. Therefore, the expectation of the bilateral linear drift and sine drift is zero, and the positive linear drift and positive sqrt drift always have positive values.

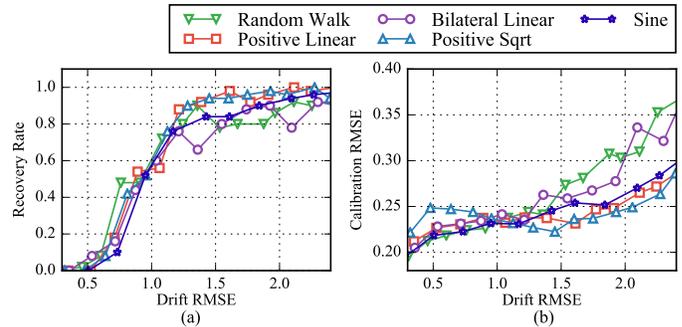

Fig. 9. (a) Recovery rates and (b) calibration RMSEs of PRNet on different types of drift under different drift levels

For each drift type and level, we run the calibration process 50 times, and for each time 20 randomly selected sensors are drifted. We plot the recovery rates and calibration RMSEs with drift RMSEs in Fig. 9. The recovery rates of different drift types show similar trends. As we identify the drifted sensors using the amplitude of the estimated drift, when the drift is small, it is more difficult to identify drifted sensors accurately. When the drift RMSE is larger than 1.2, for every drift type, PRNet can obtain a recovery rate higher than 0.6. For calibration RMSEs, when the drift RMSE is lower than 1.2, the calibration RMSEs for different drift types are lower than 0.25. The increasing trends of calibration RMSEs with drift RMSEs for different drift types are slightly different. This is caused by the different features of drifts. However, in many cases, random-walk drift, which is the same as the training data, does not have the highest recovery rate or the lowest calibration RMSE, since the randomness and uncertainty of this model make it difficult to calibrate. Therefore, we can conclude that PRNet trained with the augmented data does not overfit the random-walk drift. As long as the drift level is within a reasonable range, the sensor drift can be effectively suppressed. If the drift characteristics in a specific application is not well-simulated by our model, we could simply apply a new model in the drift-generation step to fit the application, which does not substantially change the PRNet framework.

We also test PRNet on calibrating noisy measurements with random-walk drift and sine drift. By carefully choosing the drift level parameters, the two drift models produce similar drift RMSEs, where $\tilde{\sigma}_d$ is set to 0.02 and $S$ to 5.5. For each test case, 20 random selected sensors are drifted, and all sensors are corrupted by Gaussian noise. Table II shows the calibration RMSEs on different noise levels, where the calibration RMSEs of random-walk and sine drift are very close. As we train PRNet at the noise level of $\sigma_n = 0.5$, in the test cases, when $\tilde{\sigma}_n$ is smaller than 0.5, or the signal-noise-ratio (SNR) is higher than 26 dB, the calibration RMSEs are almost the same. When it increases over 0.5, the calibration error slightly increases. During training, we should choose the noise level based on application requirements. If the training noise is too large, the calibration precision for low-noise measurements will decrease.

One requirement in training neural networks is that the training data should cover the features of the test data.

TABLE II
CALIBRATION RMSEs ON DIFFERENT NOISE LEVELS AND DRIFT TYPES

| $\tilde{\sigma}_n$ | 0 | 0.1 | 0.2 | 0.5 | 0.7 | 1.0 |
|---|---|---|---|---|---|---|
| SNR (dB) | Inf | 39.9 | 33.9 | 25.9 | 23.0 | 19.9 |
| Random Walk Drift | | | | | | |
| D.RMSE | 1.52 | 1.54 | 1.55 | 1.59 | 1.68 | 1.82 |
| C.RMSE | 0.272 | 0.261 | 0.268 | 0.265 | 0.302 | 0.333 |
| Sine Drift | | | | | | |
| D.RMSE | 1.55 | 1.53 | 1.56 | 1.64 | 1.71 | 1.88 |
| C.RMSE | 0.251 | 0.238 | 0.249 | 0.266 | 0.290 | 0.312 |

Note: D.=Drift C.=Calibrated

TABLE III
ARCHITECTURES OF NETWORKS IN BENCHMARK

| Network | PRNet | FCN | ColFCN |
|---|---|---|---|
| # of layers | 13 | 26 | 15 |
| Param-size | 334.4 KB | 698.9 KB | 300.0 KB |
| Time complexity* | $O(14208N+N^2)$ | $O(36304N)$ | $O(19724N+N^2)$ |
| Computation* | 750.0 KOps | 1.792 MOps | 966.7 KOps |
| Receptive field | (58, 15) | (56, 56) | (72, 23) |

* normalized to computation amount for each ($N$, 1) measurement

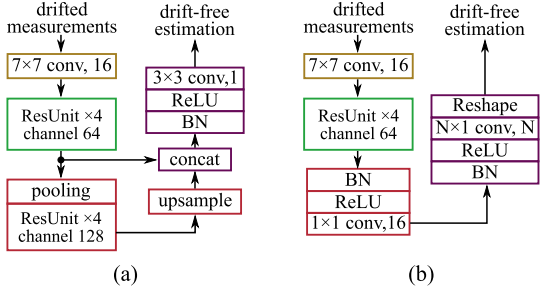

Fig. 10. The architecture of two baseline networks: (a) a multi-stage FCN model and (b) ColFCN model with a large kernel in the last layer.

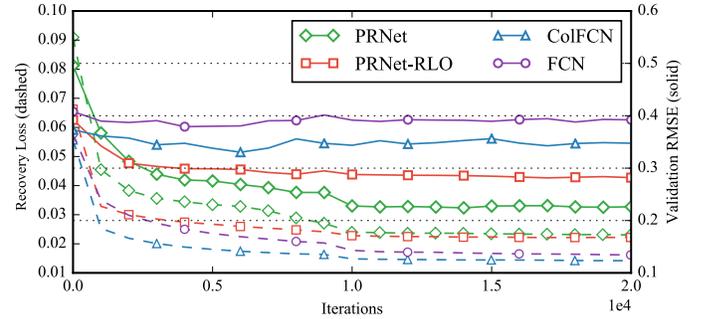

Fig. 11. Recovery losses (dashed line) and validation RMSEs (solid line) of PRNet, PRNet-RLO, ColFCN and FCN.

In this experiment, we show that random-walk is an appropriate model which approximately covers various drift types. We also show that PRNet trained with the proposed data augmentation method can calibrate noisy measurements. In real-world applications, we can also add possible corruptions such as instant pulse noise [6] to cover application specific data features.

## V. EXPLORATION ON ARCHITECTURE AND PARAMETERS

In this section, we use the simulated dataset to explore the influence of different settings to PRNet, including architecture design, temporal kernel size, training patch size, and data-rearrangement.

### A. Architecture Design

In this experiment, we show the effectiveness of PRNet architecture by comparing its performance with three other architectures, including:
- PRNet-RLO: the same architecture with PRNet, trained with recovery loss only (RLO);
- FCN: a two-stage fully convolutional network (FCN) illustrated in Fig. 10(a);
- ColFCN: an FCN with a large convolution kernel in the last layer, illustrated in Fig. 10(b). Compared with Fig. 5(a), ColFCN is similar to a "reversed" PRNet.

The numbers of layers, parameter sizes, computation time complexity, computation amounts, and the sizes of receptive fields of these networks are listed in Table III. The proposed PRNet has the smallest computation amount and a small parameter size; the ColFCN has the largest receptive field size over the space dimension, and the FCN model is the deepest and has the largest computation and parameter amount.

We do not need to run both the pre-training and fine-tuning processes to benchmark these models, since their performance gap appears in the pre-training process. Every 1000 iterations, we validate them with 100 cases of drifted measurements, in each of which 20 randomly selected sensors are drifted, and the drift parameter $\tilde{\sigma}_d^2$ is set to 0.01. The input data patch size is set to 100 to satisfy the need of the FCN model. We plot the training losses and validation RMSEs in Fig. 11. As all the networks except for PRNet do not have a projection loss, only the recovery loss is accounted.

An interesting phenomenon is that the proposed PRNet has the smallest validation RMSE, while it has the largest training loss. First we compare PRNet and PRNet-RLO. Without the projection loss, as the recovery loss becomes the only optimization objective, it is easier for PRNet-RLO to reach a lower recovery loss. However, for the projection layer, separating the information of the ground truth signal and the sensor drift gets more difficult, so PRNet-RLO has a higher validation RMSE than PRNet. Comparing the FCN and ColFCN models, the ColFCN model has both lower training loss and validation RMSE than FCN. This is because that the ColFCN model has a larger spatial receptive field than FCN, so that it can better utilize the global spatial correlations of the sensing field.

It is obvious that the projection layer has much contribution in improving the calibration accuracy. In essence, the projection layer introduces a prior assumption to the network that the sensor drift can be observed in a projected space, making it easier for the recovery layers to extract and calibrate the drift. From the network perspective, the FCN and ColFCN models actually have better representation ability since they are deeper and wider than PRNet, but without the projection layer, they overfit the training set and fail to extract essential features of the sensory data.

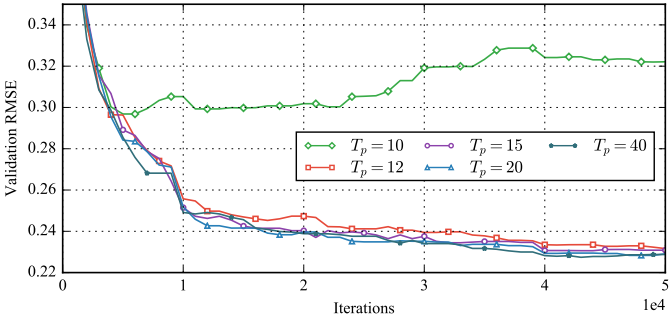

Fig. 12. Validation RMSEs of PRNets trained with different sizes of training patches.

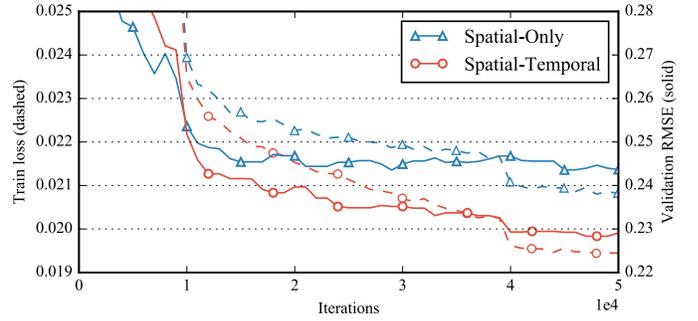

Fig. 13. Training losses (dashed line) and validation RMSEs (solid line) of spatial-temporal PRNet and spatial-only PRNet.

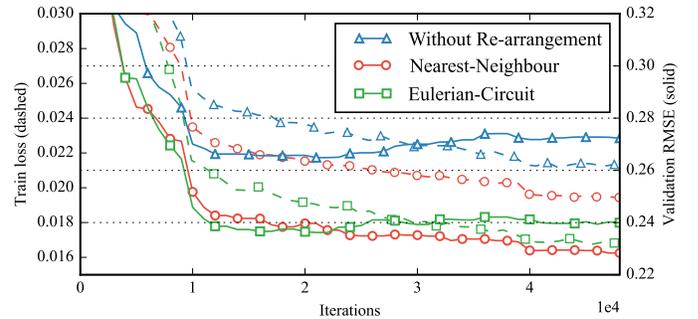

Fig. 14. Training losses (dashed line) and validation RMSEs (solid line) of PRNets using training data with or without data re-arrangement.

## B. Patch Size Selection

In Section III-D we claim that the patch size of training data should be slightly larger than the receptive field of the neural network over the time dimension. In this experiment, we test the performance on different patch sizes.

As shown in Table III, the receptive field size of PRNet over the time dimension is 15. We set the patch size $T_p$ to 10, 12, 15, 20 and 40, and train 5 PRNets. As the training loss is related to the patch size, we use the validation RMSE to compare their performances.

Fig. 12 shows the validation RMSEs of PRNets trained with different patch sizes. When the patch size is 10, the validation RMSE gets stuck at a high level, and even slightly increases with training iterations. This is because that the training patch cannot even fill the receptive field of the network, and the features of training data are quite different from those of validation and test data. In contrast, PRNets trained with other patch sizes show very similar performance in validation RMSE. The performance gain of larger patch size is quite limited. Furthermore, larger patch size leads to more computation and a slower training speed. Therefore, as we have suggested, a number that is slightly larger than the receptive field of the network is an appropriate selection. In most of our experiments, the patch size for training PRNet is 20.

## C. Benefit From Temporal Correlation

In this experiment we show how PRNet utilizes temporal correlation and how temporal correlation contributes to calibration performance. We derive a spatial-only PRNet based on the original one, where all $k_s \times k_t$ convolution kernels are replaced by $k_s \times 1$ kernels. This means that the spatial-only PRNet computes on measurements collected at a single time instant. Next we increase the channels of the spatial-only PRNet, so that it has the parameter size of 348.3KB and normalized computation amount of 1.164MOps, which are slightly larger than that of the original one.

We plot the training loss and validation RMSEs of the two PRNet models during the pre-training process in Fig. 13. Where both the training loss and validation RMSE of spatial-temporal PRNet are lower than those of spatial-only PRNet. Therefore, spatial-temporal convolution helps PRNet to utilize the temporal correlation of measurement data.

## D. Necessity of Data Re-Arrangement

In this experiment we show why sensor re-arrangement is indispensable. We train three PRNets with the same architecture using the same training data, but different re-arrangement matrices, including
- without re-arrangement: set $M$ to an identity matrix;
- nearest neighbour: calculate $M$ using nearest-neighbour algorithm;
- Eulerian circuit: re-arrange sensors by traversing over the Eulerian circuit on the edge-duplicated MST.

The training losses and validation RMSEs of these PRNets during the pre-training process are plotted in Fig. 14. The one trained without re-arrangement has the largest training loss and validation RMSE. The other two PRNets achieve similar validation RMSEs, but the one trained with Eulerian-circuit-based re-arrangement has lower training loss. As the Eulerian-circuit solution ensures that the maximum distance of neighbour sensors is minimized, the organized data should have more local correlation. However, the validation curve shows this does not bring much benefit in validation. Therefore, the sensor re-arrangement step is required, but a greedy approximated solution can provide enough local data correlation.

We plot the locations of sensors in Fig. 15, in which the sensor numbers in Fig. 15(a) are in the original order whereas those in Fig. 15(b) is re-arranged using the nearest-neighbour algorithm. Obviously, in the original order, measurements form neighbour sensors may distribute far apart in the measurement matrix, leading to the uncorrelation of adjacent rows, reducing the calibration performance of PRNet. By re-arranging the sensor order, neighbour sensors are more possible to be put

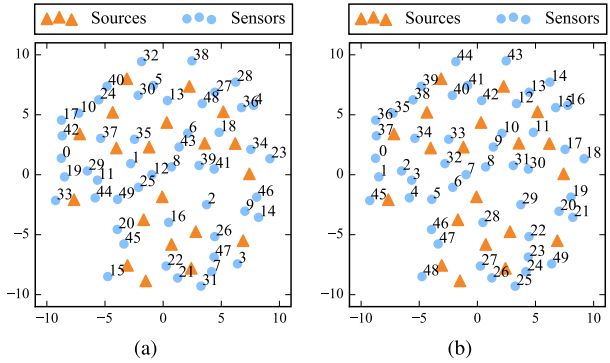

Fig. 15. Locations of signal sources and sensors in the simulated sensing field. The number beside the sensor represent its row index in the measurement matrix. (a) Original order. (b) Rearranged using greedy algorithm.

in adjacent rows, ensuring the local spatial correlation and improving the calibration performance.

## VI. Discussion

Although we had lots of empirical experiment results evaluating the effectiveness of PRNet, it is difficult to theoretically explain why PRNet surpasses existing methods. In this section, we try to give an intuitive explanation.

PRNet has two stages: projection and recovery. In Section III-B we have explained that the projection stage is a natural extension to the linear low-dimensional projection used in the subspace method, which can be applied in more applications such as nonlinear sensing fields.

For the recovery layers, we claim that the key ingredient is the nonlinear ReLU activation function, given by

$$\text{ReLU}(x) = \max(0, x). \tag{30}$$

We first consider the case that the activation functions are linear, then the overall recovery function is also linear. The optimal solution for this case will be a linear least-square regression, which does not have the ability of detecting and calibrating drifted sensors. However, with the ReLU activation function, some of the neurons' outputs are suppressed to zero, while the rest are linear combinations of the inputs. Therefore, the overall drift-estimation function can suppress the outputs corresponding to non-drifted sensors to zeros, while those of the drifted ones are segmented linear combinations of the input with the least square error. This property of multi-layer neural networks is called universal approximation [35].

The ResNet architecture, compared with conventional neural networks, has even higher representation ability. In [27], Hardt and Ma propose a reduced ResNet architecture, and prove that each ResUnit can map some of the input to arbitrary values while keeping the remaining values unchanged. By stacking these ResUnits, the ResNet can map the input to any wanted values. Although they reduce the ResUnit architecture in order to make it mathematically provable, it is reasonable that the full version of ResNet has similar properties.

In addition, as ResNet is optimization-friendly [27], [28], it is possible that by training PRNet, it can converge to a near optimal case that the calibrated measurements of non-drifted sensors keep the same with the input, while those of the drifted sensors are mapped to drift-free values.

## VII. Conclusion

With the increasing deployment of large-scale and long-term wireless sensor networks, sensor drift is becoming a serious problem, while calibrating sensors one-by-one is impractical when a sensor network can scale as large as hundreds of sensors. Blind calibration is a practical scheme that recovers drift-free sensory data from drifted measurements without the ground truth, but it is difficult to blindly calibrate general monitoring sensor networks for the lack of a prior data model and low-density deployment.

In this paper, we propose a deep learning approach to blindly calibrate sensor measurements named PRNet. We assume that the sensors are calibrated before deployment, so the measurements collected during the initial period can be treated as drift-free. Using the proposed data augmentation method to generate training data from initial measurements, we train PRNet to automatically extract spatial and temporal features of sensor measurements and suppress the drift. Experiments show that our PRNet has superior performance over previous methods in both recovery rate and calibration RMSE, and it can also calibrate noisy measurements with various types of drifts. Therefore, for long-term general monitoring sensor networks with sensors deployed in fixed locations, the proposed method can blindly calibrate sensor measurements, ensuring the data quality and validity.

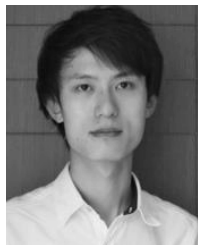

**Anqi Yang** received the B.S. degree from the Department of Electronic Engineering, Tsinghua University, Beijing, China, in 2014, where he is currently pursuing the Ph.D. degree, under the supervision of Prof. H. Yang. His current research interests include wireless sensor networks, signal processing, compressed sensing, and deep neural networks.

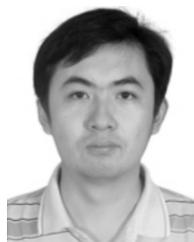

**Xiaoming Chen** (S'12–M'15) received the B.S. and Ph.D. degrees from the Department of Electronic Engineering, Tsinghua University, Beijing, China, in 2009 and 2014, respectively. He was a Post-Doctoral Research Scholar with the Electrical and Computer Engineering Department, Carnegie Mellon University, from 2014 to 2016. He is currently a Visiting Assistant Professor with the Department of Computer Science and Engineering, University of Notre Dame. His current research interests include CAD algorithms for reliable and trusted circuit design and internet of things. He received the 2015 EDAA Outstanding Dissertation Award. He also received three best paper nominations in ISLPED'09, ASP-DAC'12, and ASP-DAC'14.

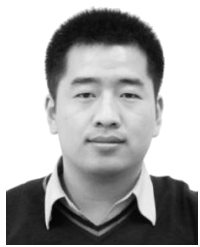

**Pengjun Wang** received the B.S. and Ph.D. degrees from the Department of Electronic Engineering, Tsinghua University, Beijing, China, in 2006 and 2011, respectively. He is currently an Assistant Research Scientist with the Department of Electronic Engineering, Tsinghua University. Since 2014, he has been the CTO of Smartbow Tech. Inc. His recent research mainly focuses on wireless sensor networks and structural health monitoring.

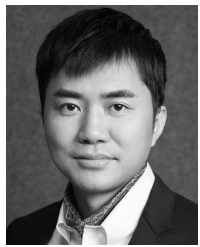

**Yu Wang** (S'05–M'07–SM'14) received the B.S. degree from Tsinghua University, China, in 2002, and the Ph.D. (Hons.) degree from the NICS Group, Electronics Engineering Department, Tsinghua University, in 2007. He is currently an Associate Professor with the Electronic Engineering Department, Tsinghua University. He is also a Founder of Deephi Tech. His research interests include brain-inspired computing system design with both CMOS and emerging devices.

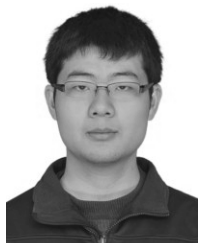

**Yuzhi Wang** (S'14) received the B.S. degree from the School of Telecommunication Engineering, Xidian University, Xi'an, China, in 2012. He is currently pursuing the Ph.D. degree with the Department of Electronic Engineering, Tsinghua University, Beijing, China, under the supervision of Prof. H. Yang. His research interests include wireless sensor networks, Internet of Things, network security, machine learning, and deep neural networks.

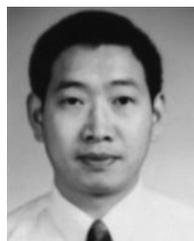

**Huazhong Yang** (M'97–SM'00) received the B.S. degree in microelectronics and the M.S. and Ph.D. degrees in circuits and systems from Tsinghua University, Beijing, China, in 1989, 1993, and 1998, respectively. Since 1993, he has been with the Department of Electronic Engineering, Tsinghua University, where he has been a Full Professor since 1998, and a Specially-Appointed Professor of the Cheung Kong Scholars Program since 2011. He has authored and co-authored over 300 technical papers. He holds 70 granted patents. His current research interests include wireless sensor networks, structural health monitoring, brain inspired computing, nonvolatile processors, and energy-harvesting circuits.